\pdfoutput=1
\documentclass{article}
\usepackage[preprint]{neurips_2026}
\usepackage[utf8]{inputenc}
\usepackage[T1]{fontenc}
\usepackage{hyperref}
\usepackage{url}
\usepackage{booktabs}
\usepackage{amsmath,amssymb}
\usepackage{graphicx}
\usepackage{microtype}

\newcommand{\dk}{d_{k}}
\newcommand{\dv}{d_{v}}
\newcommand{\dm}{d_{\mathrm{model}}}

\title{Direct access attention: associative recall at a fraction of the compute
of softmax, linear attention and state-space models}

\author{%
  Simon Dennis \quad Rivaan Patil \quad Hao Guo \quad Kevin Shabahang \\
  i14 \\
  \texttt{simon.dennis@i14.ai}
}

\begin{document}
\maketitle

\begin{abstract}
Direct access attention replaces the search over context with a learned integer lookup. A
token is projected to a $\dk$-dimensional code whose signs select one of $2^{\dk}$ memory
cells; its value is written to that cell, and a query with the same signs retrieves it by
integer index in $O(\dk{+}\dv)$ operations per head, independent of the context length $T$.
The mechanisms in current use are either expensive to decode from or limited in the number of
associations they can separate. Softmax attention compares a query with every key in the
context and returns a weighted average of the corresponding values, so emitting one token
takes $O(T(\dk{+}\dv))$ operations per head, and softmax separates $2^{O(\dk)}$ pairs, where
$\dk$ and $\dv$ are the key and value widths. Linear attention instead superposes all pairs
into a single matrix and contracts that matrix against the query in $O(\dk\dv)$ operations,
constant in $T$, but this fixed-width representation separates only $O(\dk)$ pairs per head.
Direct access gives a sum of the two widths in place of their product, and $2^{O(\dk)}$ cells
per head in place of $O(\dk)$ associations, at the cost of a table of $2^{\dk}\dv$ values per
head.

The corresponding training rule remains differentiable without constructing the
$2^{\dk}\!\times\!\dv$ table. Because the codebook contains every $\pm1$ pattern, a softmax
over its cells factorises into $\dk$ independent bits. At inference, the integer lookup is the
saturating limit of this same expression. The two coincide once an address commits to a single
cell, but cross-entropy alone does not enforce such commitment. We therefore add address entropy to
the loss, only after the task error has aligned the codes.

On in-context associative recall, one-layer models with matched parameter counts achieve
perfect accuracy at every task size from $8$ to $512$ stored pairs, both with direct
access and with softmax attention. At $512$ pairs, linear attention falls to $0.340$, while a
selective state-space model with the same state size is at chance beyond $32$ pairs. Although
both direct access and softmax hold that accuracy, their decoding costs differ by three
orders of magnitude: on the attention route, direct access requires $1{,}694$ FLOPs per
decoded token regardless of context length, against $2{,}137{,}344$ FLOPs for softmax at $512$
stored pairs ($1{,}104$ context tokens).
\end{abstract}

\section{Introduction}

\begin{figure}[t]
\centering
\includegraphics[width=0.84\textwidth]{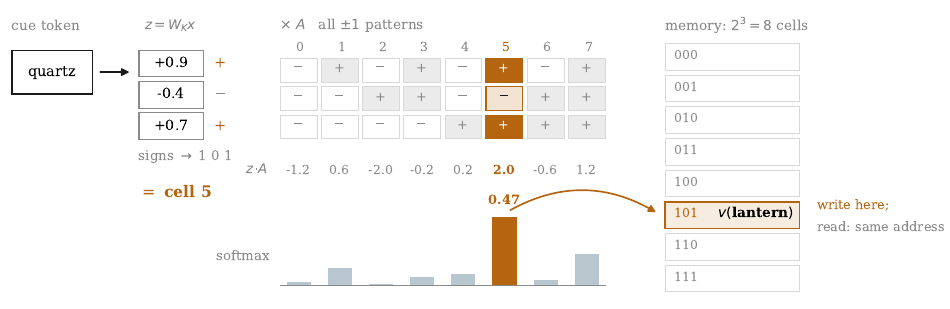}
\caption{Direct access attention at $\dk{=}3$, with real numbers. The cue's code
$z=(+0.9,-0.4,+0.7)$ has sign pattern $1\,0\,1$, which selects cell~$5$ of the eight, and the
target's value is written there and read back from it. The codebook $A$ holds every $\pm1$
pattern as a column, which is why the highest-scoring column is the one matching the signs of
$z$ (Section~\ref{sec:relax}). Training uses the softmax over cells, shown at the bottom; at
inference the cell index is computed from the sign bits alone
(Section~\ref{sec:compiled}).}
\label{fig:arch}
\end{figure}

We propose direct access attention, in which the storage location is a learned function of
the key (see Figure~\ref{fig:arch}). A cue is projected to a short real-valued vector, its
code; the sign pattern of that code selects one of $2^{\dk}$ cells, and the target's value is
added into that cell. Retrieval is then an index rather than a search, so the cost of a read
does not depend on the context length, and the number of cells available to store associations
separately is exponential in the width of the code rather than linear.

The mechanisms in current use have one of those two properties and not the other. In softmax
attention every key-value pair is retained as a separate entry, and retrieval requires scoring
the query against every stored key, so the cost of decoding one token grows with the context
length. Linear attention instead accumulates the pairs as outer products in a single
$\dk\!\times\!\dv$ matrix, from which a value is retrieved by a matrix-vector product; that
cost is constant in the context length, but the number of associations the matrix retains
before substantial interference scales as $O(\dk)$ \citep{ramsauer2021hopfield}.

Cell assignments cannot be learned by gradient descent, since the map from a code to an
integer index is piecewise constant. Training therefore replaces the hard index with a softmax
over the $2^{\dk}$ cells. That softmax factorises into $\dk$ independent per-bit factors
because the codebook contains every $\pm1$ pattern, and the sum over cells can be exchanged
with the sum over positions because the memory is a sum of outer products. The read is then a
weighted sum over the context, each weight a product of $\dk$ per-bit factors, so the cost of
training grows with the context length, as in softmax attention, but not with the number of
cells, and no quantity of size $2^{\dk}\dv$ is formed or maintained.

The relaxation of a complete $\pm1$ codebook, and the factorisation of the softmax over its
cells into independent bits, are due to \citet{zeng2023lookupffn} and are reused by
\citet{ding2024memoryformer}, to replace a dense layer with a table of trained weights, leaving
the attention operation intact. Direct access attention instead writes the values of the
current sequence into the table with key codes and reads them with query codes, and its read is
the expectation of the stored values over all $2^{\dk}$ cells rather than one row of the table
scaled by a softmax coefficient. The exchange of sums above requires that expectation
(Appendix~\ref{app:related}).

The contributions are as follows.

\begin{itemize}\itemsep2pt
\item A memory addressed by the sign pattern of a learned code decodes a token in
$O(\dk{+}\dv)$ operations per head, independent of the context length, against the
$O(\dk\dv)$ contraction of linear attention and the $O(T(\dk{+}\dv))$ search of softmax
(Section~\ref{sec:cost}).
\item A $\dk$-bit code addresses $2^{\dk}$ cells at a projection cost linear in $\dk$, so the
$22$ heads of $12$ bits used here hold $90{,}112$ associations separately, where a
$\dk\!\times\!\dv$ matrix of the same widths retains $264$ (Section~\ref{sec:cost}).
\item The weight linking a query to a value written at another position has a closed form
costing $\dk$ multiplies, so no quantity of size $2^{\dk}$ is formed at any point in training,
and the integer lookup is the saturating limit of that weight rather than a rounded version
(Section~\ref{sec:training}).
\item The entropy of an address has a closed form costing $\dk$ sigmoids per position, and
adding it to the loss drives each address onto a single cell, which is the condition under
which the integer read reproduces the differentiable one exactly. The penalty has to be delayed
until the task loss has aligned the codes: an address committed earlier is committed to an
arbitrary cell (Section~\ref{sec:commit}).
\item Five mechanisms are compared at identical parameter counts on in-context associative
recall, giving the number of stored pairs at which each one loses accuracy
(Section~\ref{sec:main} and Appendix~\ref{app:full}).
\item A budget of $24$ address bits divided among eight heads four ways, at identical
parameters and decode cost, moves the compiled read from $0.485$ to $1.000$ at $N{=}64$, the
largest effect on the compiled read of any single change reported in this paper
(Appendix~\ref{app:alloc}).
\end{itemize}

\section{The Model}\label{sec:method}

This section defines the address, the memory it indexes, and the cost and capacity that
follow from them.

The signs of a code determine which cell it selects; the magnitudes play no part in that
choice and matter only during training. Successful retrieval requires the key and query
projections to produce matching sign patterns rather than similar vectors. Throughout, $H$ is
the head count, $\dm$ the residual width, $N$ the pairs per episode, $T$ the context length and
$w$ a window width.

\subsection{The address}

Per head, a token's residual $x_t$ gives three projections, a key code, a query code and a
value:
\begin{equation}\label{eq:proj}
z_k(t)=W_K x_t,\qquad z_q(t)=W_Q x_t\ \in\mathbb{R}^{\dk},\qquad
v(t)=W_V x_t\ \in\mathbb{R}^{\dv}.
\end{equation}
The codes are real-valued; only the address read off them is binary. Taking the sign of each
coordinate as one bit,
\begin{equation}\label{eq:cell}
c(z)=\sum_{i} 2^{i}\,\mathbf{1}[z_i\ge 0]
\end{equation}
is the number of the cell a code selects, one of $2^{\dk}$.

\subsection{The memory}\label{sec:mechanism}

Each position writes its value into the cell its key code selects, and a query reads the cell
its own code selects:
\begin{equation}\label{eq:hard}
\widehat{M}_t[c]=\!\!\sum_{j\le t,\ c(z_k(j))=c}\!\! v(j),
\qquad
y_t=\frac{\widehat{M}_t\big[c(z_q(t))\big]}{\max\big(n_t[c],1\big)},
\end{equation}
with $n_t[c]$ the number of keys in cell $c$, so a cell holding several keys returns their
mean. The first line is a scatter into a table and the second an indexed gather; neither
involves a comparison with the stored keys, and neither depends on how many positions have
been written. Equations \eqref{eq:cell} and \eqref{eq:hard} are the mechanism proposed here
and the one used at inference. Section~\ref{sec:training} shows how projections that produce
matching addresses can be learned.

\subsection{Cost and capacity}\label{sec:cost}

Forming an address takes $\dk$ sign tests, and reading and normalising a cell takes $O(\dv)$ operations, so a read costs $O(\dk{+}\dv)$ per head and does
not depend on the context length $T$: a sum of the two widths, where linear attention contracts
a $\dk\!\times\!\dv$ state in $O(\dk\dv)$ and softmax scores $T$ keys in $O(T(\dk{+}\dv))$.
And a $\dk$-bit code addresses $2^{\dk}$ cells while costing $\dk$ rows each of $W_K$ and
$W_Q$, so the number of associations a head keeps separate is exponential in a width whose
parameter cost is linear in it, against the $O(\dk)$ a $\dk\!\times\!\dv$ matrix retains
before substantial interference \citep{ramsauer2021hopfield}. At the widths used here, $22$
heads of $12$ bits address $90{,}112$ cells where linear attention separates $264$ pairs.

The cost appears instead in storage. A dense table holds $2^{\dk}\dv$ values and $2^{\dk}$ counts per
head, against the $O(T(\dk{+}\dv))$ cache of softmax and the $O(\dk\dv)$ state of linear
attention, so the exponential term removed from the read reappears in the memory.
Table~\ref{tab:cost} gives all three, with analytic decode counts, alongside the other
mechanisms compared in Section~\ref{sec:main}.

\section{Training the Addresses}\label{sec:training}

The map \eqref{eq:cell} from a real-valued code to an integer cell index is piecewise
constant, so differentiating the read of \eqref{eq:hard} gives nothing that would move a key
or a query to a different cell. Training the mechanism of Section~\ref{sec:method} therefore
requires, in turn, a relaxation of the address that has gradients, an evaluation of that relaxation which never forms the
$2^{\dk}$ cells, a normalisation that keeps an unoccupied address from returning a confident
value, the condition under which the relaxed read and \eqref{eq:hard} coincide, and a loss
term that brings the relaxation to that condition.

\subsection{The hard address provides no gradient}\label{sec:relax}

Fix $A\in\{-1,+1\}^{\dk\times 2^{\dk}}$ whose column $c$ is the $\pm1$ sign pattern of cell
$c$. $A$ enumerates \emph{every} pattern. In place of the single cell that \eqref{eq:cell}
selects, training uses a distribution over all $2^{\dk}$ of them,

\begin{equation}\label{eq:addr}
a(z)=\mathrm{softmax}\big(z^{\!\top}\!A\big).
\end{equation}

The complete $\pm1$ codebook has two properties the addressing relies on. The logit of cell $c$ is
$\sum_j z_jA_{jc}$, a sum in which each bit appears once, so it is maximised by choosing
$A_{jc}=\operatorname{sign}(z_j)$ separately for every $j$, which is possible precisely because
$A$ enumerates all $2^{\dk}$ patterns.\footnote{Equivalently: take any cell disagreeing with
$\operatorname{sign}(z)$ on bit $j$. Completeness puts the cell that agrees with it elsewhere
and matches $\operatorname{sign}(z_j)$ in the codebook too, and that cell scores $2|z_j|$
higher, so repeating the swap ends at the sign-matching cell. Under an incomplete codebook the
corrected competitor need not exist, and one large coordinate could carry a cell with the wrong
sign elsewhere.} Hence the winning cell is the one whose column of $A$ matches the signs of $z$
entry by entry. Its logit is $\sum_j|z_j|$, and the cell that differs from it in bit $j$ alone
scores $2|z_j|$ less, so increasing $|z_j|$ widens that margin without affecting the optimal
value of any other bit.

The memory of \eqref{eq:hard} becomes the expectation of its write and its read under that
distribution:
\begin{equation}\label{eq:mem}
\begin{aligned}
M_t[c] \;&=\; \sum_{j\le t} a\big(z_k(j)\big)_c\,v(j)\ \in\mathbb{R}^{\dv}
  \qquad\text{for each of the } 2^{\dk}\text{ cells } c,\\[0.3em]
\tilde y_t \;&=\; \sum_{c=0}^{2^{\dk}-1} a\big(z_q(t)\big)_c\,M_t[c]\ \in\mathbb{R}^{\dv}.
\end{aligned}
\end{equation}
The first line gives the contents of one cell as a sum over the positions written so far; the
second reads by summing over all $2^{\dk}$ cells. A position adds its value into every cell, in
the proportion its address assigns to that cell, and a read combines the contents of every
cell, in the proportion the query's address assigns. Written as matrices the two lines are
$M_t=\sum_{j\le t} a(z_k(j))\otimes v(j)\in\mathbb{R}^{2^{\dk}\times\dv}$ and
$\tilde y_t = a(z_q(t))^{\!\top}M_t$. $M$ is the relaxed counterpart of $\widehat M$, and the
write and the read remain independent operations on the same table.

\subsection{The table must not be constructed}\label{sec:fact}

Implemented as \eqref{eq:mem} is written, the training arithmetic would grow exponentially
with the address width: the differentiable read forms a $2^{\dk}$-wide outer product at every
position, on top of the $2^{\dk}\dv$ numbers per head the table itself holds. 
This is
wasteful even at widths where the table fits in memory: at $\dk{=}12$, $\dv{=}32$ it reserves
$131{,}072$ numbers per head to hold at most $T{=}1104$ written values, and at $\dk{=}24$ it
would be $537$ million.

The table is therefore not constructed during training. Appendix~\ref{app:derivation} removes
the cell axis in six steps, each written out at $\dk{=}2$ where the sum over cells has four
terms and can be given in full. The derivation establishes three results, and the rest of the
paper uses only those.

First, because the codebook is complete, the softmax over cells factorises into $\dk$
independent bits, and the inner product of two addresses collapses the same way into a product
of per-bit agreement probabilities:
\begin{equation}\label{eq:kernel}
\kappa_{tj}\;=\;\big\langle a(z_q(t)),\,a(z_k(j))\big\rangle
\;=\;\prod_{i=1}^{\dk}\frac{1+\tanh\!\big(z_q(t)_i\big)
                             \tanh\!\big(z_k(j)_i\big)}{2}.
\end{equation}
This kernel is the probability that a cell drawn from the query's distribution equals one drawn
from the key's, which is the probability that all $\dk$ bits agree. It costs $\dk$ multiplies
and never forms the cell axis. The $\tanh$ follows directly from writing the per-bit agreement
probability in terms of the code, and its range is the whole of $(-1,1)$ because $z$ is
unbounded, so a bit with a large-magnitude code coordinate contributes a factor approaching
$1$ when the two codes agree and approaching $0$ when they do not.

Second, the read itself needs no cell axis. $M_t$ is a sum of outer products, so the sum over
cells in \eqref{eq:mem} can be exchanged with the sum over positions, leaving
$\tilde y_t=\sum_{j\le t}\kappa_{tj}\,v(j)$: a sum over the $t$ positions written so far rather
than over $2^{\dk}$ cells. Both transformations are necessary and serve different
purposes: exchanging the two sums removes the need to construct the table, and the
factorisation makes the result cheap, since evaluating each $\kappa_{tj}$ from its definition
would itself be a $2^{\dk}$-term inner product.

Third, because the bits are independent, the entropy of an address is a sum of $\dk$ binary
entropies,
\begin{equation}\label{eq:ent}
H\big(a(z)\big)=\sum_{j=1}^{\dk} H_b\big(\sigma(2 z_j)\big),
\qquad
H_b(p)=-p\ln p-(1{-}p)\ln(1{-}p),
\end{equation}
in nats, with $\sigma(x)=1/(1+e^{-x})$ the logistic function. It runs over the whole range:
$\dk\ln2$ when every bit is undecided, and $0$ in the limit of large $|z_j|$, where the address
is a single cell. Both ends are reachable because $z$ is unbounded, and
Section~\ref{sec:commit} describes how training is made to reach the low-entropy end.

\subsection{An unoccupied address must not return a confident value}\label{sec:divisor}

Naive normalisation can turn an unoccupied address into an apparently confident answer. Write
$\mathrm{tot}_t=\sum_{j\le t}\kappa_{tj}$ for the total kernel mass at the query's
address. At an unoccupied address, which is a miss, the kernel weights are of order
$10^{-30}$, and dividing them by their own sum rescales them to sum to one, so the read
returns a weighted average of the residual contributions from non-matching keys. Cross-entropy
cannot then distinguish a hit from a miss, because the two produce outputs of the same
magnitude.

Imposing a lower bound on the divisor prevents this rescaling. Dividing instead by $\max(\mathrm{tot}_t,\kappa_{\max})$ for a
fixed $\kappa_{\max}$ leaves a genuine hit unchanged, since its mass is far above the floor,
and collapses a miss towards zero, since its mass is far below it. The floor used here is
$\kappa_{\max}=\big((1+\tanh^{2}1)/2\big)^{\dk}$, the value \eqref{eq:kernel} takes for two
codes that agree on every bit at unit magnitude, which is $0.0591$ at $\dk{=}12$, and the read
is
\begin{equation}\label{eq:read}
y_t=\frac{\tilde y_t}{\max\!\big(\mathrm{tot}_t,\ \kappa_{\max}\big)}
   =\frac{\sum_{j\le t}\kappa_{tj}\,v(j)}{\max\!\big(\mathrm{tot}_t,\ \kappa_{\max}\big)}.
\end{equation}
The read is not sensitive to the precise value of this constant: once the codes are
committed a genuine match scores close to $1$ and a query on an empty cell close to $0$, so
any floor between them separates the two.

\subsection{The relaxed read and the indexed read coincide at saturation}\label{sec:compiled}

After training the differentiable read of \eqref{eq:read} is replaced by the indexed read of
\eqref{eq:hard}. Nothing is retrained: the same $W_K$, $W_Q$ and $W_V$ are used, and unlike
$M$ the table $\widehat M$ is built explicitly. The compiled read is the saturation limit of
the differentiable read, and the equivalence follows in three steps.

\emph{The limit of the kernel.} As $|z_i|\!\to\!\infty$ every $\tanh$ in \eqref{eq:kernel}
tends to $\pm1$, so each per-bit factor tends to $1$ where the two codes agree and to $0$ where
they disagree. The kernel $\kappa_{tj}$ therefore becomes the indicator that the two codes
agree on every bit, which is the indicator that the key and the query select the same cell.

\emph{The limit of the normalised read.} Each part of \eqref{eq:read} then collapses onto its
counterpart in \eqref{eq:hard}. Its numerator $\sum_j \kappa_{tj}v(j)$ becomes the sum of
$v(j)$ over the keys in the query's cell, which is $\widehat M_t[c]$. Its divisor
$\mathrm{tot}_t=\sum_j\kappa_{tj}$ becomes a count of those same keys, which is $n_t[c]$. And
the floor $\kappa_{\max}$ plays the part of the $\max(\cdot,1)$, so that in both cases a query on
an empty cell divides zero by a positive number. Averaging rather than summing is therefore not a
choice made for the compiled path; it is the normalisation of \eqref{eq:read} read at
saturation.

\emph{The two reads coincide.} The writes agree as well, since a committed $a(z_k)$ puts its
mass in the cell $c(z_k)$ selects, which is the write $\widehat M$ receives. At zero address
entropy, then, \eqref{eq:read} and \eqref{eq:hard} are the same function of the same weights:
the substitution is exact in the limit rather than approximate. Equation \eqref{eq:entloss}
supplies the regularisation needed to approach that limit during training;
Section~\ref{sec:commit} defines it and gives the reason it has to be scheduled.

\subsection{Commitment must follow alignment}\label{sec:commit}

The two reads coincide only when an address has committed to a single cell, and the task loss
alone does not enforce this. The cross-entropy term reaches a low value long before the address commits,
so training stops with $a(z)$ spread over many cells. Equation \eqref{eq:hard} keeps only the
largest cell, and the contributions of the remaining cells are discarded at substitution.

Two separate conditions have to hold for an indexed read to work, and each is supplied by a
different part of the objective.

\textbf{Coincidence.} The query and its key must select the same cell. This can only come from
the task error, since only a wrong value at a retrieval position signals that two codes
disagree. No auxiliary term carries information about which key corresponds to a given query.

\textbf{Commitment.} The address must put its mass on one cell, so that discarding the
remaining cells leaves the read unchanged. Cross-entropy does not enforce this, since a spread address that returns the
right answer by averaging reaches the same loss value, and an auxiliary penalty is therefore
required. By \eqref{eq:ent} the entropy of an address is available in closed form, so it can
be added to the loss directly:
\begin{equation}\label{eq:entloss}
\mathcal{L}=\mathcal{L}_{\mathrm{CE}}
 \;+\;\beta\sum_{\text{heads}}\Big[\overline{H\big(a(z_k)\big)}+\overline{H\big(a(z_q)\big)}\Big],
\end{equation}
averaged over positions, on keys and queries alike. Because \eqref{eq:ent} needs no cell
axis, the term costs $\dk$ sigmoids per position and nothing else.

The two requirements have to be met in that order, and the term is scheduled accordingly. An
address driven onto a single cell before the task loss has aligned the codes is a committed
address on an arbitrary cell, and once every bit is saturated no gradient remains to move it. The term is
therefore \emph{activated on convergence}: it is switched on at the first evaluation where the
differentiable read passes a threshold accuracy, and stays on from then. The weight $\beta$
is set to $1$; Appendix~\ref{app:commit} shows accuracy is unchanged for any $\beta$ between
$0.3$ and $30$, so its value does not have to be tuned.

\section{Experimental Setup}\label{sec:setup}

This section gives the task and the reason
for choosing it, the models, the training settings and how accuracy is measured.

\paragraph{Task.} The task is in-context associative recall. Each trial presents $N$
cue-target pairs once, then $64$ distractors drawn from the same entity vocabulary so they
cannot be filtered by identity, then $8$ questions, each showing a cue whose target the model
must emit. A trial at $N{=}3$ reads

\begin{center}
{\small\texttt{otter violin\quad cedar anchor\quad quartz lantern\quad SEP\quad
$\langle64$ distractors$\rangle$\quad quartz MASK}}
\end{center}

\noindent where the spacing is for readability only: the pairs are adjacent tokens with no
delimiter between them, and \texttt{SEP} and \texttt{MASK} are the only tokens in the stream
that are not drawn from the entity pool. The correct prediction at the mask is
\texttt{lantern}, from a binding seen once, $67$ tokens earlier, among five other entities
from the same pool.

Bindings are uniform over $|V|{=}512$ entities and redrawn every trial, so nothing can be
memorised and accurate prediction requires retrieval from the current context. There is consequently no
fixed test set: evaluation draws fresh trials on a separate random stream, and generalisation
means generalisation to new random bindings rather than to held-out items. Three choices in the
generator keep the task a test of retrieval. Cues within a trial are distinct, so every question has exactly one right
answer. Questions are sampled with replacement, because drawing them without it makes the
question set a permutation of the facts and admits a process-of-elimination strategy
yielding $0.34$ accuracy at $N{=}8$ with no retrieval. Targets are masked in the input while remaining the
training target, so that a repeated question cannot be answered by copying a nearby answer
instead of retrieving the distant fact. The episode is $2N{+}81$ tokens and the model input is
that minus one, so $T=2N{+}80$: $144$ at $N{=}32$, $208$ at $N{=}64$, $592$ at $N{=}256$ and
$1104$ at $N{=}512$.

\paragraph{Why this task is diagnostic.} Raising $N$ tests how many associations a
mechanism can keep apart, and the three mechanisms with capacity bounds reach that limit at very different points.
Linear attention superposes its pairs into one matrix and holds $O(H\dk)$ of them before
crosstalk, $264$ at the widths used here, where direct access attention has $O(H2^{\dk})$
cells, $90{,}112$: linear against exponential scaling in the same width. Softmax retains each
pair separately, with a bound of $O(\min(T,e^{cH\dk}))$, likewise above the largest
task size swept here. Sweeping from $8$ to $512$ stored pairs therefore crosses one of the three bounds
and not the other two, so linear attention should be the only one of the three to lose
accuracy in the range. That is also
why the paper's two claims are made against different baselines: capacity against linear
attention, whose limit falls inside the swept range, and computational cost against softmax,
whose does not.

The selective SSM has no bound of that form, which is why the capacity column of
Table~\ref{tab:cost} is marked \textsc{n/a} for it, but its state can be matched: at the conventional
$d_{\mathrm{st}}{=}16$ it holds $d_{\mathrm{in}}d_{\mathrm{st}}=528\times16=8{,}448$ numbers,
the same as linear attention's $H$ matrices of $\dk\!\times\!\dv$, $22\times12\times32$. The
sweep puts the two at identical state size and measures the task size at which each loses
accuracy.

\paragraph{Input representation.} Each position receives the concatenated embeddings of the
last two tokens, $[\,\mathrm{emb}(x_{t-1});\mathrm{emb}(x_t)\,]$, drawn from one shared table
of width $\dm/2$, so that a position carries a whole cue-target pair rather than half of one.
This makes a one-layer comparison possible. With one token per
position no position holds both halves of a binding, so a causal model must copy the cue
forward and then match against it, and \citet{sanford2024onelayer} show that solving this
induction-head task with one layer requires exponentially greater size than with two.
Supplying the pair at one position removes that circuit, so the comparison reported here varies
capacity rather than depth.

\paragraph{Models.} Every model is a single transformer layer of width $\dm{=}264$ with $22$
heads and an MLP of four times the residual width. A head carries $\dv{=}32$ value dimensions
and $\dk{=}12$ key dimensions, which for the direct-access arm means $4{,}096$ cells per head.

The layer is implemented with each of five mechanisms in turn. Full softmax attention is the upper baseline, since
it keeps every pair and can score a query against all of them. Sliding-window softmax at
$w{=}8$ is the lower one, since its reach is shorter than the gap between a fact and the
question that asks for it. Linear attention, the constant-cost superposition, is the baseline for the
address-capacity claim. The selective state-space model is the other constant-state
baseline, following Mamba at its conventional state size, $d_{\mathrm{st}}{=}16$ over
$d_{\mathrm{in}}{=}528$ channels, and specified in full in Appendix~\ref{app:repro}. The fifth is
direct access attention.

The arms are matched on parameter count, so that a difference in accuracy between them cannot
be attributed to model size. Matching first requires a common notion of head width: $\dk$ is the
address width in the direct-access arm and $d_{\mathrm{head}}$ the query-key width in the
others, and the two are set equal. Each of the four attention-style arms then has
$\dm(2\sum_h \dk^h + 2H\dv)$ parameters, which is $511{,}104$ on the attention route and
$1{,}735{,}932$ in the whole model at $N{=}256$, the positional table growing with $T$. The
selective SSM has no head structure, so that expression does not apply to it: at
$d_{\mathrm{st}}{=}16$ it carries $464{,}640$ parameters in its mixer, and at the
$d_{\mathrm{st}}{=}48$ reported alongside it in Section~\ref{sec:main} it carries
$515{,}328$. The two settings bracket the common count rather than matching it, which is why
both are reported.

Appendix~\ref{app:alloc} departs from these settings, and varies the direct-access arm alone. It
uses $8$ heads rather than $22$, and a budget of $\sum_h\dk^h{=}24$ address bits rather than
$264$, divided among those heads in seven different ways, at $N{=}32$ and $N{=}64$. The
smaller budget produces measurable failures inside the range tested, and the smaller task
sizes make a full sweep of the partitions computationally feasible.

\paragraph{Cost and capacity.} Table~\ref{tab:cost} gives, for each of the five mechanisms,
the number of pairs it can separate, its asymptotic decode cost and that cost evaluated at two
context lengths, for the attention route alone with the state or cache already built.

\begin{table}[t]
\centering\small
\caption{Asymptotics and analytic decode cost, per token, for the attention route alone, at
$H{=}22$, $\dk{=}12$, $\dv{=}32$ and window $w{=}8$, with the selective SSM at
$d_{\mathrm{in}}{=}528$ and $d_{\mathrm{st}}{=}16$. Parameters are $2H\dm(\dk{+}\dv)$ for the
four attention-style rows; the SSM row carries $9\%$ fewer, and has no standard closed-form
address-capacity bound.}
\label{tab:cost}
\setlength{\tabcolsep}{3pt}\begin{tabular}{llllrr}
\toprule
mechanism & address capacity & decode FLOPs & state & $T{=}112$ & $T{=}65{,}536$\\
\midrule
softmax                & $O(\min(T,e^{cH\dk}))$ & $O(TH(\dk{+}\dv))$ & $O(TH(\dk{+}\dv))$
  & $216{,}832$ & $126{,}877{,}696$\\
sliding window        & $O(w)$ & $O(wH(\dk{+}\dv))$ & $O(wH(\dk{+}\dv))$
  & $15{,}488$ & $15{,}488$\\
linear attention       & $O(H\dk)$ & $O(H\dk\dv)$ & $O(H\dk\dv)$
  & $16{,}896$ & $16{,}896$\\
selective SSM          & \textsc{n/a} & $O(d_{\mathrm{in}}d_{\mathrm{st}})$
  & $O(d_{\mathrm{in}}d_{\mathrm{st}})$ & $54{,}912$ & $54{,}912$\\
direct access attention & $O(H2^{\dk})$ & $\mathbf{O(H(\dk{+}\dv))}$
  & $O(H2^{\dk}\dv)$ & $\mathbf{1{,}694}$ & $\mathbf{1{,}694}$\\
\bottomrule
\end{tabular}
\end{table}

Every attention-style mechanism in the table carries the same $2H\dm(\dk{+}\dv)$ parameters,
so the differences in decode cost and capacity among them follow only from where the pairs are
stored. The contrast with linear attention is
direct. A superposed state has to be contracted against the query in full, $\dk\dv$
multiply-accumulates, a \emph{product} of the two widths. An addressed state is indexed
instead: $\dk$ sign tests that form an integer, a gather of $\dv$ floats from the cell that
integer indexes, and $\dv$ divisions by the cell's occupancy count, which is
$\dk{+}2\dv{+}1$ operations, a \emph{sum} of the two widths where the contraction is a
product of them. At $\dk{=}12$ and $\dv{=}32$ that is $77$ operations per head against $768$.

\paragraph{Commitment.} The address entropy term of \eqref{eq:entloss} is applied to keys and
queries alike with weight $\beta{=}1$, activated at the first evaluation where the differentiable
read exceeds $0.9$ accuracy and left on thereafter. Appendix~\ref{app:commit} reports its
sensitivity.

\paragraph{Training.} Every run trains for $30{,}000$ steps at batch $64$ with AdamW, at a
learning rate of $10^{-3}$ with $(\beta_1,\beta_2)=(0.9,0.95)$, decayed on a cosine schedule after $400$
warmup steps. Weight decay is $0.1$ and applies to matrices only, gradients are clipped at
$1.0$, the seed is $0$ and the model is evaluated every $2500$ steps.

Two implementation choices reduce memory use without changing the computed function: the
differentiable read is evaluated in chunks of $128$ query positions, and gradient
checkpointing is on. Together they allow the $O(T^{2})$ read to run at $T{=}1104$.
Both are defaults of the training loop.

\paragraph{Metric.} Accuracy is exact match on the single token that answers each question, and
only that position is scored: the trainer also supervises the step before it, the same retrieval
one token earlier, but does not report it. Each evaluation draws $8$ batches of $64$
episodes with $8$ questions, so every number is a mean over $4096$ answer tokens, and chance
is $1/512=0.002$. Addressing arms are reported through the \textbf{compiled read} unless the
differentiable read is named, and both reads are evaluated on the same batch with the same weights in
one forward pass. Appendix~\ref{app:repro} gives the remaining reproduction detail.

\section{Results}\label{sec:main}

\begin{figure}[t]
\centering
\includegraphics[width=0.88\textwidth]{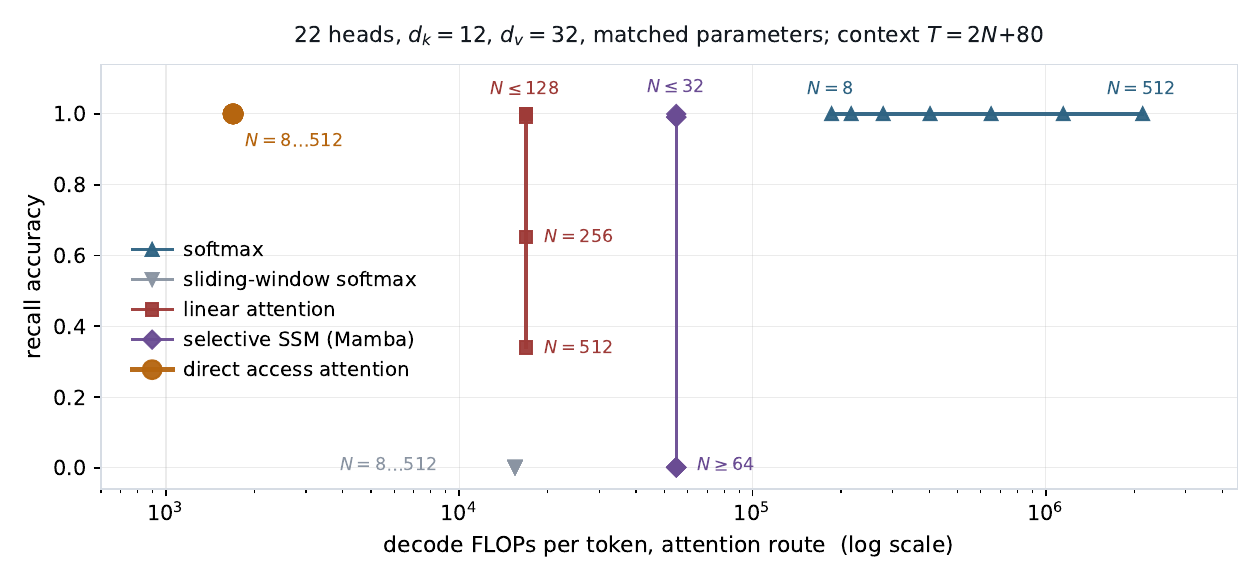}
\caption{Recall accuracy against the cost of emitting one token, for five mechanisms at
matched parameter counts. Each point is one task size, the stored pairs running from $8$ to
$512$ along every series with the context growing with them, $T{=}2N{+}80$. Accuracy is the
compiled read for direct access attention and the ordinary read otherwise; cost is the
analytic FLOP count for the attention route alone, given in Table~\ref{tab:cost}. The
selective SSM is at its conventional state size, $9\%$ short of the other four on parameters;
a run at the larger state size that matches them is reported in the
text, and Appendix~\ref{app:full} tabulates both axes.}
\label{fig:frontier_main}
\end{figure}

Figure~\ref{fig:frontier_main} plots recall accuracy against decode cost for all five
mechanisms at each task size, and Table~\ref{tab:main} gives the same accuracies as numbers.
Direct access attention's compiled read reaches $1.000$ at every $N$ from $8$ to $512$, and at
each of those sizes equals the differentiable read to three decimals: replacing the kernel of
\eqref{eq:kernel} by the integer path of \eqref{eq:hard} leaves the answers unchanged to the
reported precision.

Linear attention holds $1.000$ to $N{=}64$ and then falls, to $0.993$, $0.652$ and $0.340$ at
$N{=}128$, $256$ and $512$. Every pair written into a $\dk\!\times\!\dv$ state is added to
the same matrix, so a read recovers a target only to the extent that the other pairs'
contributions cancel, and the number of pairs that can be kept apart this way grows as
$O(H\dk)$, which is $264$ at these widths. Linear attention writes an outer product at every
token position, not only at the $N$ positions holding a fact, so the count to compare against
$264$ is $T=2N{+}80$ rather than $N$. At $N{=}64$ that is $208$, below the bound, and accuracy
is $1.000$; at $N{=}128$ it is $336$, above it, and accuracy begins to fall. That is the onset
the superposition bound predicts.

The selective SSM is the weaker of the two constant-state baselines. It solves the task to
$N{=}32$, at $1.000$, $0.991$ and $0.991$, and is at chance from $N{=}64$ onward, where linear
attention still reads $1.000$; the two carry the same state, $8{,}448$ numbers each. Widening
the state to $48$ triples the SSM's own state, which then matches the other arms on
parameters. This raises $N{=}64$ to $0.613$, but from $N{=}128$ the arm is again at
chance. Tripling the state therefore delays failure by only one doubling of $N$, at $156{,}288$
decode FLOPs against $54{,}912$, so state size alone does not explain the SSM's loss of accuracy.
Appendix~\ref{app:related} develops a structural interpretation, that a selective recurrence lacks an explicit binding operation, and relates it
to prior results: \citet{jelassi2024repeat} prove that a two-layer transformer can copy strings
whose length grows exponentially with its size while any model carrying a fixed-size latent
state cannot, and \citet{waleffe2024mambastudy} find that $8$B Mamba models match transformers
on most benchmarks but fall behind on those that require copying or in-context retrieval.

Softmax solves the task at every tested size because it keeps every pair separately, but its
decode cost grows with the context. Sliding-window attention serves as a local-access control rather than a competitive baseline:
one layer of an eight-token window reaches seven tokens back, against a gap of at least $66$
tokens, so the facts are outside its receptive field by construction. It remains at chance throughout, confirming that the task cannot be solved from
the question's immediate neighbourhood.

In Figure~\ref{fig:frontier_main} softmax maintains constant accuracy at a decode cost that
grows with the context, a horizontal line; linear attention and the selective SSM have
constant decode cost and falling accuracy, vertical ones. All direct-access points coincide,
at the lowest cost of the five: at $N{=}512$ the compiled read requires $1/1{,}262$
of softmax's decode FLOPs and a tenth of linear attention's.

How a fixed budget of $24$ address bits is divided among $8$ heads moves the compiled
read from $0.485$ to $1.000$ at $N{=}64$, at identical parameters and identical decode cost,
though not at an identical table size: the best division allocates sixteen times as many cells
as the uniform one, since the cell count grows as $\sum_h 2^{\dk^h}$ where the parameter count
grows as $\sum_h \dk^h$.
Neither the uniform division nor the most concentrated one is best: two heads of nine bits
with a tail of one-bit heads reads $1.000$, against $0.485$ for eight equal heads of three at
the same parameters and decode cost, and $0.002$ for a single head of $24$ bits. Divisions
that change the head count, as that last one does, span the wider range but differ in
parameters and decode cost. These
experiments do not establish why. Width is not limiting at the sizes tested: at $22$ heads and
$N{=}128$, a trained model at $\dk{=}12$ writes to only $59$ of its $4{,}096$ cells per head,
and models trained at $\dk{=}6$, $8$ and $10$ all read $1.000$ on both paths.
Appendix~\ref{app:alloc} reports the allocation sweep in full.

The advantage claimed here is at decode, not training. The differentiable read is
$O(T^{2}(\dk{+}\dv))$, the same order as softmax, and slower in practice because the kernel accumulates
$\dk$ per-bit factors instead of issuing one matrix multiplication: a training step at
$T{=}112$ and batch $64$ takes $17.9$\,ms against $3.4$\,ms for softmax.
Appendix~\ref{app:repro} gives the remaining timings and a faster kernel that the saturated
codes would permit, which is not implemented here.

\section{Conclusion}

Softmax attention keeps every key-value pair as a separate cached entry; linear attention
adds it into a single superposition. In neither case does the storage location depend on what
the pair contains, so retrieval requires a comparison against every stored key or a
contraction of the superposition rather than an integer lookup. Making the storage location a
learned function of the key turns a read into an index. The addressed memories cited above build the
table they index and fill it with weights trained over a corpus. Here a complete $\pm1$
codebook makes the softmax over cells factorise, and associativity exchanges the sum over
cells for a sum over positions, so no table is formed in training. The integer read used at
inference is the saturating limit of the differentiable read rather than a surrogate fitted
to it.

On a retrieval task storing $8$ to $512$ pairs per episode, the compiled read scores $1.000$
at every size for $1{,}694$ FLOPs per decoded token. Softmax reaches the same accuracy but
requires up to $1{,}262$ times as many FLOPs on the attention route, its cost growing with the
context. The two constant-cost baselines require fewer FLOPs than softmax but lose accuracy as the
number of stored associations grows: linear attention falls to $0.340$ at $512$ pairs at ten
times the decode cost of the compiled read, and a selective state-space model is at chance from $64$ pairs. The integer read agrees with the differentiable read it replaces to three
decimals at every size, so the substitution causes no measurable loss.

\label{endofmain}%
\bibliographystyle{plainnat}
\bibliography{references}

\appendix

\section{Removing the cell axis}\label{app:derivation}

The six steps below remove every reference to the $2^{\dk}$ cells of \eqref{eq:mem} from the
quantities training uses. Each is written out at $\dk{=}2$, where the
sum over cells has four terms and can be given in full.

\paragraph{Step 1. The partition function factorises.}
The numerator of \eqref{eq:addr} is already a product over bits, since the exponential of a
sum is a product of exponentials:
$\exp(\sum_j z_jA_{jc})=\prod_j\exp(z_jA_{jc})$. It remains to show that the denominator
factorises as well. At $\dk{=}2$ the four cells give
\begin{equation}\label{eq:zfac}
\underbrace{e^{z_1+z_2} + e^{z_1-z_2} + e^{-z_1+z_2}
  + e^{-z_1-z_2}}_{\text{four cells}}
\;=\;
\underbrace{\big(e^{z_1}+e^{-z_1}\big)
            \big(e^{z_2}+e^{-z_2}\big)}_{\text{two bits}},
\end{equation}
and multiplying the right-hand side out returns the four terms on the left. In general
$Z=\prod_j 2\cosh(z_j)$, a product of $\dk$ terms in place of a sum of $2^{\dk}$.
\citet{zeng2023lookupffn} derive the same identity, and \citet{ding2024memoryformer} reuse
it; both apply it to normalise a scalar coefficient attached to a hard lookup into a table that
is materialised. The steps below carry the factorisation through the read as well, so that the
cell axis is removed from the read and not only from the normaliser.

The derivations below depend on this step, and it holds \emph{only} because the columns of
$A$ enumerate all $2^{\dk}$ sign patterns. That is exactly the condition for the exponents on
the left to run over every combination of $\pm z_j$, which allows them to be collected into
the product on the right. For a codebook that is not complete, $Z$ does not factor and
none of the steps below apply.

\paragraph{Step 2. The address factorises into $\dk$ independent bits.}
Both the numerator and the denominator of \eqref{eq:addr} are now products over bits, so the
quotient is a product of per-bit quotients:
\begin{equation}\label{eq:perbit}
a(z)_c=\frac{\prod_j e^{z_jA_{jc}}}{\prod_j\big(e^{z_j}+e^{-z_j}\big)}
      =\prod_{j=1}^{\dk}\frac{e^{z_jA_{jc}}}{e^{z_j}+e^{-z_j}}.
\end{equation}
Each factor depends on one bit of cell $c$ and one coordinate of $z$. Since $A_{jc}$ is $+1$ or
$-1$, dividing numerator and denominator by $e^{z_jA_{jc}}$ leaves $1/(1+e^{-2z_jA_{jc}})$,
which is the logistic function $\sigma(x)=1/(1+e^{-x})$ evaluated at $2z_jA_{jc}$. Hence
\begin{equation}\label{eq:bern}
a(z)_c=\prod_{j=1}^{\dk}\sigma\big(2 z_j A_{jc}\big).
\end{equation}

Read as a probability distribution, \eqref{eq:bern} says that the address is $\dk$ independent
coin tosses, one per bit, with the $j$th coming up $+1$ with probability
\begin{equation}\label{eq:pj}
p_j \;=\; \Pr[\text{bit } j = +1] \;=\; \sigma(2 z_j),
\end{equation}
since $\sigma(2z_j)+\sigma(-2z_j)=1$ and the probability of a cell is the product of the
probabilities of its $\dk$ bits. The sign of $z_j$ determines the more probable value of the
$j$th bit and $|z_j|$ determines its probability: $p_j$ is $\tfrac12$ at $z_j{=}0$ and approaches $1$ or $0$ as $z_j$
grows in either direction. At $\dk{=}12$ a distribution over $4{,}096$ cells is therefore
described by twelve numbers, and nothing of width $2^{\dk}$ has to be represented to hold it.

\paragraph{Step 3. The overlap between two addresses factorises as well.}
Write the query's per-bit probabilities as $p_j$ and the key's as $r_j$. At $\dk{=}2$ the inner product is a
sum over four cells, and collecting the two terms that agree on bit~$1$ and then the two
that agree on bit~$2$ turns it into two factors:
\begin{align}\label{eq:collapse}
\langle a(q),a(k)\rangle
&= (1{-}p_1)(1{-}p_2)(1{-}r_1)(1{-}r_2) + (1{-}p_1)p_2(1{-}r_1)r_2 \notag\\
&\qquad + p_1(1{-}p_2)r_1(1{-}r_2) + p_1p_2r_1r_2 \notag\\[0.3em]
&= \big[(1{-}p_1)(1{-}r_1)+p_1r_1\big]\,\big[(1{-}p_2)(1{-}r_2)+p_2r_2\big].
\end{align}
The four-term sum has become two factors; at $\dk{=}12$, $4{,}096$ terms become twelve. Each bracket is
the probability that its bit agrees between the two addresses.

\paragraph{Step 4. The kernel, in closed form.}
The identity $2\sigma(2x)-1=\tanh(x)$ rewrites \eqref{eq:pj} as
$p_j=\tfrac12(1{+}\tanh z_{q,j})$, and likewise $r_j=\tfrac12(1{+}\tanh z_{k,j})$ for the key.
Substituting both into \eqref{eq:collapse}, each bracket becomes
$(1{+}\tanh z_{q,j}\tanh z_{k,j})/2$, and the product of the brackets is the kernel
\eqref{eq:kernel} quoted in Section~\ref{sec:fact}. This is the
property the compiled read of Section~\ref{sec:compiled} relies on and the loss term of
Section~\ref{sec:commit} produces: as $|z_i|$ grows, $\kappa_{tj}$ approaches the indicator
that every bit agrees, which is the quantity the compiled read computes.

\paragraph{Step 5. The read can be evaluated without the cell axis.}
Steps 1 to 4 make one address cheap to describe, but the read in \eqref{eq:mem} still
contracts against a $2^{\dk}\!\times\!\dv$ table. Associativity, rather than factorisation,
removes that contraction: $M_t$ is a sum of outer products, so the contraction can be taken in the other
order,
\begin{align}\label{eq:reassoc}
\tilde y_t\;=\;\sum_{c=0}^{2^{\dk}-1} a\big(z_q(t)\big)_c\,M_t[c]
&\;=\;\sum_{c=0}^{2^{\dk}-1} a\big(z_q(t)\big)_c \sum_{j\le t} a\big(z_k(j)\big)_c\,v(j)
\notag\\[0.3em]
&\;=\;\sum_{j\le t}\bigg(\underbrace{\sum_{c=0}^{2^{\dk}-1}
      a\big(z_q(t)\big)_c\,a\big(z_k(j)\big)_c}_{\textstyle\kappa_{tj}}\bigg)v(j)
\;=\;\sum_{j\le t}\kappa_{tj}\,v(j).
\end{align}
Substituting the definition of $M_t$ leaves the sum over cells outside the sum over positions;
exchanging the two puts it inside, where it is exactly the inner product of two addresses,
which \eqref{eq:kernel} evaluates in $\dk$ multiplies. The cell axis is summed away rather than
traversed: the first line runs over $2^{\dk}$ cells and the last over the $t$ positions written
so far, and nothing of width $2^{\dk}$ is ever allocated on the differentiable path. Both
operations are necessary and serve different purposes: associativity removes the need to
construct the table, and factorisation makes the result cheap, since evaluating $\kappa_{tj}$
from its definition would itself be a $2^{\dk}$-term inner product, leaving the exponential
cost in the kernel.

\paragraph{Step 6. The entropy factorises as well.}
The entropy of a set of independent variables is the sum of their entropies, so the entropy of
an address is a sum of $\dk$ binary entropies, one per bit, each evaluated at the $p_j$ of
\eqref{eq:pj}. That is \eqref{eq:ent}, and like the other quantities here it is computable
without the cell axis.

\section{Reproduction detail}\label{app:repro}

\paragraph{Code layout.} Every file named in this appendix is in the released
repository. The mechanisms are in \texttt{binary\_memory/lm.py}, the task
generator in \texttt{binary\_memory/facts.py} and the training loop in
\texttt{binary\_memory/train\_lm.py}, whose \texttt{DEFAULTS} dictionary holds every
hyper-parameter not overridden on the command line. \texttt{compare.py} defines every arm
reported here, in its \texttt{BASE} string and \texttt{ARMS} dictionary, and is the entry
point that produced them. \texttt{flops.py} computes Table~\ref{tab:cost} through its
\texttt{analytic\_decode} function, and \texttt{bench\_step.py} measures the training-step
times quoted in Section~\ref{sec:main}, on one GPU at the headline shape with forward,
backward and optimiser step included. Four scripts read saved checkpoints: \texttt{align.py} gives
the cell occupancy and agreement figures of Appendix~\ref{app:alloc}, \texttt{dead\_bits.py}
the per-bit balance and distinct-cell counts behind the address-width figures,
\texttt{union\_growth.py} the saturation of the cell union over episodes, and
\texttt{probe\_common.py} holds the model reconstruction all of them share, so that every
probe rebuilds the model exactly as it was trained.

\paragraph{Why $w{=}8$ is a local-access control.} Over $L$ layers a window of $w$ relays information
$L(w-1)$ tokens, so one layer at $w{=}8$ reaches $7$. The nearest fact sits
$\mathrm{gap}+2 = 66$ tokens before the first question. The sliding-window arm therefore
cannot see the facts and serves as a local-access control rather than a competitive baseline.
\texttt{train\_lm.py} asserts $L(w-1) < \mathrm{gap}+2$ before training starts, so training
terminates if a configuration would put a fact inside the window.

\paragraph{Selecting the variants.} \texttt{compare.py} builds its job list from
\texttt{BASE} plus per-arm overrides. Head count is set by the \texttt{CMP\_HEADS}
environment variable (22 by default, 8 for the reduced-budget results) and the entity pool
by \texttt{CMP\_ENT}, which is $512$ throughout. Mixed address
widths are given by a \texttt{d\_k\_mix} string of the form \texttt{9x2,1x6}, meaning two
heads of width~9 followed by six of width~1; \texttt{train\_lm.py} expands it to a per-head
list and \texttt{lm.py} groups heads of equal width, giving each group its own $W_K$ and
$W_Q$ while $W_V$ and the output projection stay shared.

\paragraph{The linear-attention baseline.} The linear arm uses the
$\mathrm{elu}(x){+}1$ feature map of \citet{katharopoulos2020linear}, so a read is
$\phi(q)^{\!\top}\!\sum_{j\le t}\phi(k_j)v_j^{\!\top}$ normalised by
$\phi(q)^{\!\top}\!\sum_{j\le t}\phi(k_j)$, with $\phi$ applied to the same $\dk$-wide
projections every other arm uses.

\paragraph{The selective SSM baseline.} \texttt{MambaBlock} in \texttt{lm.py} is a
from-scratch S6 block, not a wrapper around the released kernels, so that it runs in the same
container as everything else. It follows \citet{gu2023mamba}: an input projection to
$d_{\mathrm{in}}{=}2\dm{=}528$ channels and a gate of the same width, a causal depthwise
convolution with $d_{\mathrm{conv}}{=}4$ taps, SiLU, input-dependent $B$, $C$ and $\Delta$
with $\Delta$ passed through softplus from a low-rank projection initialised so that the
initial time constants span $0.001$ to $0.1$, a diagonal $A$ initialised to
$-1,\dots,-d_{\mathrm{st}}$, and the state read out as $y_t=\sum_n C_{t,n}h_{t,n}$ followed
by a gated RMSNorm before the output projection. The recurrence is evaluated by the chunked
scan of the same file, which forms the pairwise decay $\exp((S_t-S_s)A)$ from the cumulative
sum of $\Delta$ inside a chunk and carries the chunk boundary state forward; without the
RMSNorm the state grows without bound over a $1{,}104$-token context and training diverges.
The state size is $d_{\mathrm{st}}{=}16$ for the reported runs and $48$ for the follow-up of
Section~\ref{sec:main}, which puts $464{,}640$ and $515{,}328$ parameters in the block against
the $511{,}104$ the other arms spend; task, optimiser and schedule are the shared ones. Being
a reference implementation, it is also the slowest arm to train: $40.3$\,ms per step at
$T{=}112$ and batch $64$ on an H200, against $14.4$\,ms for direct access and $3.8$\,ms for
linear attention (\texttt{bench\_step.py}).

\paragraph{The address probe.} The occupancy and agreement figures quoted in
Appendix~\ref{app:alloc} come from \texttt{align.py}, which
loads a saved checkpoint, regenerates validation episodes, and for each question locates
the fact position holding its answer by matching the question's cue token against the fact
cues, which are distinct within an episode. It then compares $\operatorname{sign}(z_q)$ at
the question against $\operatorname{sign}(z_k)$ at that fact position, over $4$ batches of
$8$ episodes.

\paragraph{Compute.} Runs are single-GPU, on NVIDIA H100 80GB and H200 141GB nodes. Cost is
dominated by the $O(T^{2})$ differentiable read, so it scales with $N$: the $N{=}32$ and $N{=}64$
allocation runs take about $6$ minutes each, while $N{=}512$ takes $54$ minutes for softmax,
$64$ for linear attention, $193$ for the selective SSM and $848$ for direct access. Taken together the runs reported here total
roughly $100$ GPU-hours, with a median of $16$ minutes each. Several runs share a node, so these are elapsed
times rather than exclusive-GPU times. Per step, measured on one H200 at $T{=}112$ with batch
$64$, direct access takes $17.9$\,ms with gradient checkpointing, on in every direct-access run
here, and $14.4$\,ms without it, against $3.8$\,ms for linear attention and $3.4$\,ms for
softmax, whose fused kernel is faster at a context this short. A faster path is available but
not implemented here: once the codes saturate, every $\tanh$ in \eqref{eq:kernel} is $\pm1$, so
the product of per-bit factors is the indicator that all $\dk$ bits agree, which holds exactly
when $\operatorname{sign}(z_q)\!\cdot\!\operatorname{sign}(z_k)=\dk$. The kernel could then be
computed by one matrix multiplication over the sign vectors and a comparison against $\dk$.

\paragraph{Availability.} The repository contains the mechanisms, the task generator, the
training loop, the sweep definitions and the analysis scripts named above, together with the
training logs under \texttt{logs/}. Every run directory holds the resolved configuration in
\texttt{config.json}, the full evaluation trace in \texttt{log.csv} and the final metrics in
\texttt{final.json}, which is the file every number in this paper was checked against.

\paragraph{Run names.} Each reported number is associated with a named run directory under
\texttt{runs/}, so a reader can map a table cell to a job. Names carry a prefix for the
experiment and then the setting: \texttt{main\_} for the mechanism comparison, \texttt{alloc\_}
for the allocation sweep, \texttt{width\_} for the address-width sweep, \texttt{mamba16\_} and \texttt{mamba\_} for
the selective state-space baseline at $d_{\mathrm{st}}{=}16$ and $48$, and
\texttt{commit\_} for the entropy-term ablation.

Table~\ref{tab:main} is \texttt{main\_\{softmax,window,addr\}\_N\{8,\dots,512\}}, with the
linear-attention row \texttt{main\_linear\_N*}. Table~\ref{tab:alloc} is
\texttt{alloc\_N\{32,64\}\_} followed by the allocation in row order, \texttt{24x1},
\texttt{8x3}, \texttt{4x4}, \texttt{4x5}, \texttt{2x9}, \texttt{2x12}, \texttt{1x24}.
The address-width runs are \texttt{width\_N128\_w\{6,8,10\}}, with the $\dk{=}12$ setting
shared with Table~\ref{tab:main}, and \texttt{width\_N512\_w8}. The two recoveries reported with
Table~\ref{tab:alloc} are \texttt{wd05\_N64\_c2x12}, weight decay $0.5$ in place of $0.1$, and
\texttt{lg\_N32\_c1x24}, $90{,}000$ steps in place of $30{,}000$; the two further seeds of
$8{\times}3$ at $N{=}32$ are \texttt{alloc\_N32\_8x3\_s\{1,2\}}. Appendix~\ref{app:commit} is \texttt{commit\_w*} for the
weight sweep and \texttt{commit\_s*} for the schedule.

\paragraph{Figures.} Figure~\ref{fig:arch} is drawn by \texttt{paper/fig\_arch.py} and
Figure~\ref{fig:frontier_main} by \texttt{notes/frontier\_main.py}, which calls
\texttt{analytic\_decode} in \texttt{flops.py} directly.

\paragraph{Task design.} Section~\ref{sec:setup} names the three choices in the generator
that keep the task a test of retrieval. Cues are drawn without replacement from the entity pool, which is what
makes a question's answer unique. The validation generator's seed is offset by $7777$ from the
training one.

\paragraph{The largest usable address width.} The differentiable path places no bound on
$\dk$: every quantity used in training costs $\dk$ operations per position, so doubling the
width doubles the work and still forms nothing of size $2^{\dk}$. The maximum usable width is
set by the compiled side instead. A dense table is $2^{\dk}\dv$ values per head, and the cell number of \eqref{eq:cell} is
packed into a signed $64$-bit integer by \texttt{bits\_to\_int} in
\texttt{binary\_memory/codebook.py}, which caps the width at $62$ bits whatever memory is
available.

\section{Capacity and optimisation limits of direct access attention}\label{app:alloc}

The capacity figures in Section~\ref{sec:setup} are bounds on the addresses rather than on
retrieval end to end. A retrieved value has only $\dv$ dimensions in which to represent its
target, and that is a separate limit, reached by enlarging the entity pool rather than by
storing more pairs. The sweep holds $|V|$ fixed and varies $N$, so it measures address
capacity, and the exponential claim should be read as a claim about addresses.

Section~\ref{sec:commit} identified two requirements for an indexed read, and they fail
separately. \emph{Coincidence} is whether a query selects the cell its key wrote to; it
determines whether the answer is right, and it is visible in the differentiable read.
\emph{Commitment} is whether an address has put its mass on one cell; it determines whether
compiling changes the answer, and it is visible only in the gap between the two reads. Commitment is measured here as
\emph{address concentration}, $1-H(a(z))/\dk\ln2$: the address entropy of \eqref{eq:ent} as a
fraction of its maximum, subtracted from one, so $0$ is an address spread evenly over all
$2^{\dk}$ cells and $1$ is an address on a single cell. It is reported separately for the key
and query sides, since either can be the one that fails. The sweep divides a fixed budget of address
bits among heads in different ways, and every failure it produces can be
classified as a failure of one or the other.

At a fixed head count the parameter count depends on $\sum_h \dk^h$, the \emph{total} number
of address bits, and not on how they are divided, so a budget of bits can be partitioned across
a fixed number of heads at no change in size or decode cost. A division that changes the head
count moves both, since the value path scales with $H\dv$. Write
an allocation as $n_1{\times}w_1 + n_2{\times}w_2$ for $n_1$ heads of width $w_1$ and $n_2$ of
width $w_2$; every allocation below sums to $24$ bits.

Holding parameters and decode cost fixed does not hold the table fixed. Parameters depend on
$\sum_h \dk^h$ and the compiled read of \eqref{eq:hard} costs $H(\dk{+}2\dv{+}1)$ operations
whatever the widths, but a head of width $\dk^h$ allocates $2^{\dk^h}$ cells, holding $\dv$
values and one count each, so the table grows as $\sum_h 2^{\dk^h}$. Across the four
allocations at $H{=}8$ that total is $64$, $80$, $136$ and $1{,}036$ cells: a factor of
sixteen between the uniform division and the best one, $8$\,KB against $134$\,KB of
\texttt{float32} table per layer, at identical parameters and identical decode cost. The
divisions that change the head count go further, $1{\times}24$ requiring $2^{24}$ cells and a
$2.1$\,GB table per layer. Address width therefore consumes inference
memory rather than parameters or decode operations. Both limits bound the sweep from
opposite sides: three bits per head leave too few cells to separate the pairs, and the two
widest allocations do not train.

\begin{table}[t]
\centering\small
\caption{The same $24$ address bits shared out seven ways, at two task sizes. \emph{Soft} is
the differentiable read of \eqref{eq:read}, \emph{compiled} the integer read of
\eqref{eq:hard}. \emph{Cells} is $\sum_h 2^{\dk^h}$, the size of the compiled table summed over
heads, which varies by a factor of sixteen across the four allocations at $H{=}8$ even though
their parameters and decode cost are identical. $\dv{=}32$ throughout; parameters are identical
except where $H$ differs, since the value path scales with $H\dv$. One layer, $30{,}000$ steps,
chance $1/512$.}
\label{tab:alloc}
\begin{tabular}{lccccccc}
\toprule
& & & & \multicolumn{2}{c}{$N{=}32$} & \multicolumn{2}{c}{$N{=}64$}\\
\cmidrule(lr){5-6}\cmidrule(lr){7-8}
allocation & $H$ & cells/head & cells & soft & compiled & soft & compiled\\
\midrule
$24 \times 1$                & 24 & 2 & 48 & 0.052 & 0.052 & 0.032 & 0.032\\
$8 \times 3$                 & 8 & 8 & 64 & 0.983 & 0.901 & 0.912 & 0.485\\
$4{\times}4 + 4{\times}2$    & 8 & 16 / 4 & 80 & 0.995 & 0.994 & 0.948 & 0.899\\
$4{\times}5 + 4{\times}1$    & 8 & 32 / 2 & 136 & 1.000 & 1.000 & 0.991 & 0.987\\
$2{\times}9 + 6{\times}1$    & 8 & 512 / 2 & 1{,}036 & 1.000 & 1.000 & 1.000 & 1.000\\
$2 \times 12$                & 2 & 4{,}096 & 8{,}192 & 0.992 & 0.992 & 0.004 & 0.004\\
$1 \times 24$                & 1 & 16.8M & 16.8M & 0.024 & 0.003 & 0.002 & 0.002\\
\bottomrule
\end{tabular}
\end{table}

The seven rows of Table~\ref{tab:alloc} differ only in how those bits are divided, and the
four at $H{=}8$ are matched in parameters and decode cost as well. They span the whole range of
the compiled read, from $0.002$ to $1.000$, which is the largest effect on the compiled read of
any single change reported in this paper. The best partition is neither the uniform split nor
the most concentrated: two heads of nine bits with a tail of six one-bit heads reads $1.000$ on
both reads at both task sizes, where eight equal heads of three bits, using the same budget
at the same decode cost, read $0.485$ at $N{=}64$.

\textbf{Failures of coincidence.} $2{\times}12$ solves $N{=}32$ on both reads, at $0.992$, but
at $N{=}64$ it does not rise above chance, finishing at $0.004$ on the differentiable read;
$1{\times}24$ fails at both sizes, at $0.024$ and $0.002$. Equally low accuracy on both reads
is the signature of a coincidence failure: there is no correct differentiable behaviour for
the indexed read to reproduce.
The cause is redundancy across heads rather than capacity within one. Each head forms its own
address from its own projection and the read sums over heads, so a head whose code settles on
the wrong sign in one bit selects the wrong cell, and its contribution to the read is unrelated to the target. Among twenty-two
heads the remaining twenty-one can compensate; among two, an incorrect address affects a much
larger fraction of the available heads. The difficulty is in the optimisation rather than in the capacity: raising weight
decay from $0.1$ to $0.5$ takes $2{\times}12$ at $N{=}64$ to $1.000$ on both reads, and
$1{\times}24$ at $N{=}32$ reaches $1.000$ on the differentiable read given three times the step
budget, though its compiled read then stops at $0.764$.

\textbf{Failures of commitment.} For $8{\times}3$ at $N{=}64$ the differentiable read reaches
$0.912$ but the compiled read only $0.485$, the largest disagreement between the two reads in
Table~\ref{tab:alloc} and not the only one: $4{\times}4{+}4{\times}2$ reads $0.948$ against
$0.899$ at the same task size and $4{\times}5{+}4{\times}1$ $0.991$ against $0.987$. Across these
runs the gap is ordered by concentration, and not by occupancy or per-head agreement. Either
side can be the one that falls short, the query side at $8{\times}3$, $N{=}32$ and the key side
at $8{\times}3$, $N{=}64$, so the statistic is the lower of the two. Among the allocations that
learn the task, that figure runs in order with the gap: at $N{=}64$, $0.9471$ gives $0.427$,
$0.9942$ gives $0.049$, $0.9992$ gives $0.004$ and $1.0000$ gives $0.000$; at $N{=}32$,
$0.9936$ gives $0.081$ and $0.9998$ or above gives at most $0.0001$. The same holds at
$22{\times}12$ for $N{=}128$, $256$ and $512$, where both sides reach $1.000$ and the two reads
agree to three decimals. The ordering is not a calibration: two further seeds of $8{\times}3$ at
$N{=}32$ both reach $0.9951$ and give gaps of $0.154$ and $0.053$, so concentration predicts
the scale of the gap rather than its value.
As Section~\ref{sec:compiled} predicts, the two reads diverge only where concentration stays
below one: at zero address entropy they are the same function of the same weights, so any
difference between them measures how far from that limit training stopped.

Both reads perform poorly under $24{\times}1$. Two cells per head provide too little address
resolution for this task. Three bits at $N{=}64$ produces a less extreme version of
the same address-resolution problem: eight cells per head
against $208$ positions writing into them leaves every cell holding many values, so a query
cannot do better than return their average. Commitment is incompatible with answering by
averaging, since committing such an address to one of its eight cells would discard most of
the contributions that made the answer right.

Cell occupancy, the number of keys sharing the cell a query selects, might be expected to
govern the gap, since a shared cell could make a lookup return a wrong value. It does not: the
configurations with no gap span occupancies from $0.12$ to $49.3$, and the best-performing
allocation has the most crowded cells of any run reported here. Per-head agreement between a
query and its key, the fraction of heads in which the query's cell is the cell its key wrote
to, does not govern it either. At $22{\times}12$, $N{=}512$ that fraction is $0.046$, a mean of
fewer than one head in twenty selecting the cell its key wrote to, and the model still
compiles at $1.000$. The heads are not independent: $W_K$ and $W_Q$ are trained jointly across
all heads against one loss, so there is no reason for retrieval to be spread evenly over them.
A mean of $0.046$ is consistent with a small number of heads that agree on nearly every
question, and the read requires only those heads to select the right cell.

\section{Extended related work}\label{app:related}

This appendix relates direct access attention to sparse key-value memories, efficient attention and
binary-addressed memories, then considers the recall properties of the baselines used in
Section~\ref{sec:main}.

\paragraph{Sparse key-value memory.} The most closely related work augments a transformer
with a large key-value store read sparsely. Product-key memories \citep{lample2019pkm} factor the key space
so $C$ sub-keys address $C^{2}$ slots, retrieving by top-$k$ in each half, and the idea has
since been scaled, sparsified and routed to a million single-neuron experts
\citep{kim2020largepkm,berges2024memorylayers,he2024peer,huang2025ultramem}. A parallel line
addresses a large embedding table by $N$-gram context rather than by a projected query:
deterministically from token identity in \citet{cheng2026engram}, and from discrete symbols
learned in latent space in \citet{zheng2026lngram}. These are \emph{parametric} stores read
position-wise: the slots hold trained weights and no content moves between positions through
them. The store here is filled at run time and mixes across positions.
\citet{zhao2026fwpkm} narrows that distinction, rewriting product-key slots by gradient descent
at inference so the store holds context rather than trained knowledge; a write here is a
forward-pass outer product and requires no inner optimisation. \citet{geva2021ffn} argue the feed-forward layer is already such a memory, making
the boundary one of access pattern. The role of factorisation also differs: a product key
factors the \emph{search} and still retrieves by nearest neighbour, where this factors the
\emph{softmax} into $\dk$ bits, so retrieval is one gather, with no top-$k$ hyper-parameter.

\paragraph{Retrieval augmentation and cache compression.} Two other lines cut the cost
of a softmax read by bounding how many items it scores: retrieval augmentation, which fetches
neighbours from a corpus built before the sequence is seen
\citep{khandelwal2020knnlm,borgeaud2022retro}, and cache compression, which keeps the
mechanism and discards part of the state \citep{zhang2023h2o,xiao2024streamingllm}. In both a
read is still a comparison against each item kept. The store here is filled from the context
at run time, as a cache is, but a read is an index, so its cost does not depend on how many
items the store holds and no eviction is needed to keep that true.

\paragraph{Hashing in attention.} LSH \citep{indyk1998lsh,charikar2002simhash} appears in
transformers mainly as a way to \emph{avoid} full attention, by bucketing queries and keys and
attending within a bucket \citep{kitaev2020reformer,chen2024magicpig,tay2020survey}. The hash is a fixed random
projection in that line, but it need not be: \citet{gong2025hata} and \citet{li2025spotlight}
fit binary codes for queries and keys to the attention scores of a trained model, the second
with a non-linear hash because queries and keys lie in narrow and nearly orthogonal cones. In
all of them the hash approximates an attention pattern defined independently of it, and is
measured against that pattern. The address here has no such target: it determines the read, and
the only measure of it is whether the answer is recalled. \citet{roller2021hashlayers} make the
same separation on the routing side, assigning tokens to experts by a fixed hash with no
routing parameters at all. Reformer also ties query and key
projections, because a bucketing scheme must place a query in its key's bucket. The same
constraint arises here from the other direction: with independent projections a query selects
its own key's cell essentially never at initialisation.
\citet{gulcehre2017dntm} found discrete addressing converged faster than continuous in an NTM,
and \citet{rae2019bloom} meta-learn a hashing memory.

\paragraph{Binary addresses.} Addressing by binary codes and Hamming proximity is Kanerva's
sparse distributed memory \citep{kanerva1988sdm}, related to attention by
\citet{bricken2021attnsdm} and made differentiable by \citet{wu2018kanervamachine}. SDM
scatters $M\!\ll\!2^{d}$ fixed random locations and activates every one inside a Hamming ball;
here the projection is learned, the codebook is complete, and the compiled read activates
exactly one cell. Weightless networks
\citep{bledsoe1959ntuple,aleksander1984wisard,aleksander2009weightless} use binary tuples as
literal RAM addresses but assign them by hand, and semantic hashing
\citep{salakhutdinov2009semantic} learns codes that index a static collection rather than
acting as a read and write operator. The nearest of these to the mechanism described here are
\citet{zeng2023lookupffn} and \citet{ding2024memoryformer}, which replace a dense layer with
lookup tables addressed by the sign bits of the input, score the $2^{\tau}$ cells of a
complete $\pm1$ codebook by inner product with the code, and factorise the partition function
exactly as \eqref{eq:zfac} does; the first replaces the feed-forward network and the second the
query, key and value projections, and the second cites the first. Both learn the projection
rather than sampling it. In both the read is a hard lookup of one row, with the softmax
supplying a scalar coefficient that carries gradient to the input, so the table is materialised,
and $\tau$ is held at $8$ in the second. Because the rows are parameters, a gradient and an
optimizer state of the same shape accompany the table, which \citet{zeng2023lookupffn} note as a
memory requirement of the approach, and each input supplies gradient to one row, a sparsity
\citet{ding2024memoryformer} compensate for with a learning rate three times the baseline. Their tables hold trained weights read by the code of
the position that addressed them, so no content moves between positions and the attention
operation is left untouched; \citet{ding2024memoryformer} name it as the computation that
remains after their substitution. Carrying the factorisation through the read as well, and exchanging
the sum over cells with the sum over positions, removes the table and leaves $\dk$ bounded by
the compiled side rather than by storage. Differentiable memories
\citep{graves2014ntm,graves2016dnc,rae2016sam,sukhbaatar2015memnn} learn where to read and
write but address by \emph{similarity} to stored content, and hard discrete access otherwise
needs an estimator: REINFORCE \citep{zaremba2015rlntm}, or the straight-through estimator
\citep{bengio2013ste} as used for the codebook lookup in VQ-VAE \citep{vandenoord2017vqvae}.
Neither is needed here, because the discrete read is the zero-temperature limit of the
differentiable one rather than a rounding step placed inside it. Because the integer is read off sign
bits, the compiled read is $O(\dk{+}\dv)$ rather than $O(M(\dk{+}\dv))$ over $M$ slots, and
backpropagation stays exact.

\paragraph{Recall and state size.} \citet{arora2023zoology} introduced multi-query associative
recall as a diagnostic and \citet{arora2024based} identified the tradeoff these results sit
inside, that fixed-state recurrent models struggle at recall. Linear attention
\citep{katharopoulos2020linear} superposes every pair into one $\dk\!\times\!\dv$ state, which
by the standard outer-product argument \citep{ramsauer2021hopfield} holds $O(\dk)$ pairs per
head before crosstalk. The measurements of Section~\ref{sec:main} agree, and add that
direct access is \emph{exponential} in the same width.

For selective state-space models the corresponding result is already known.
\citet{jelassi2024repeat} prove that a two-layer transformer can copy strings of length
exponential in its size while any model with a fixed-size latent state cannot, and show the
gap holds for pretrained models at scale; \citet{waleffe2024mambastudy} find that $8$B Mamba
and Mamba-2 models match transformers on most benchmarks but fall behind on those requiring
copying or in-context retrieval, and that interleaving a small fraction of attention layers
closes the gap. Section~\ref{sec:main} reproduced that ordering at a scale small enough to
isolate it, on a task that is nothing but retrieval; the next paragraph gives the structural
reason to expect it.

\paragraph{Correspondence with classical models of associative memory.} Each of the three retrieval baselines
corresponds to a classical model of human associative memory, and the correspondence
identifies which retrieval operation each architecture implements directly, and therefore
which of them must learn an approximation to an operation the others provide.

Softmax attention keeps every pair separately and answers a query by scoring it against all of
them, weighting by similarity and summing. That is MINERVA 2 \citep{hintzman1984minerva,
hintzman1986schema}: traces retained individually, a probe compared to each, and an echo
formed by similarity-weighted summation, with the softmax's exponential weighting playing the
role of Hintzman's cubing of similarity. Linear attention superposes the pairs into
$\sum_j v_jk_j^{\!\top}$ and retrieves by a matrix-vector product, which is the correlation
matrix memory \citep{anderson1972assoc,kohonen1972correlation} and the matrix model of
\citet{humphreys1989matrix}. Its $O(\dk)$ capacity before crosstalk is the classical result
for that store rather than a property of the deep-learning setting.

The selective state-space model has no comparably direct implementation. Unrolling its recurrence gives
$y_t = \sum_{s\le t}\big(C_t\!\cdot\!B_s\big)\,\gamma_{ts}\,u_s$ for a decay $\gamma$, which is
bilinear in the same way a linear attention is and can be written as one \citep{dao2024ssd};
its binding term is the single scalar $C_t\!\cdot\!B_s$, shared across every channel of the
state, rather than $H$ independent cue-target pairings, as Section~\ref{sec:main} sets out.
The resulting store is closer to the single composite trace of TODAM \citep{murdock1982todam},
in which every association is convolved into one vector and recall is an approximate
deconvolution, than to a matrix model in which each pairing has its own outer product.
Associative retrieval must emerge from the learned recurrent dynamics rather than being
carried out by an operation the architecture provides.

Direct access attention differs from all of these in using an explicit address: the cue
determines a location rather than being compared against stored content, which places it
closer to Kanerva's sparse distributed memory \citep{kanerva1988sdm} than to any of the models
above.

\paragraph{Recurrent models that overwrite.} DeltaNet \citep{schlag2021fwp,yang2024deltanet}
reads what is stored at a key before writing, so a repeated key is overwritten rather than
accumulated, and Gated DeltaNet \citep{yang2025gateddeltanet} adds Mamba2-style decay
\citep{gu2023mamba}; both beat plain linear attention at recall, and \citet{arora2024readtwice}
show how much of the remaining gap is a data-order effect. The distinction is the destination
of the write: the delta rule writes to a content key inside the same
$\dk\!\times\!\dv$ superposition, so capacity stays $O(H\dk)$ and the gain is interference
removed rather than slots added. The two could be combined, since a delta update could govern
a cell's contents. No DeltaNet baseline was run here, and it is the most important baseline
missing from the comparison.

\section{Full results by task size}\label{app:full}

Table~\ref{tab:main} gives the accuracy axis of Figure~\ref{fig:frontier_main} and
Table~\ref{tab:mainflops} the cost axis; the figure plots one against the other.

\begin{table}[t]
\centering\small
\caption{Recall accuracy against the number of stored pairs. Every mechanism but the SSM has
the same parameter count, $\dm(2\sum_h \dk^h + 2H\dv)$, which is $511{,}104$ attention
parameters and $1{,}735{,}932$ in the whole model at $N{=}256$; the SSM row carries $464{,}640$
in its mixer. One layer, $22$ heads, $\dk$ and $\dv$ widths of
$12$ and $32$, $30{,}000$ steps. Chance is $1/512$, the size of the entity pool the answer
is drawn from. The selective SSM row is at $d_{\mathrm{st}}{=}16$. The direct-access row is the
\emph{compiled} read of \eqref{eq:hard}; the other four mechanisms have no compiled read and
are reported as they run.}
\label{tab:main}
\begin{tabular}{lccccccc}
\toprule
& \multicolumn{7}{c}{number of pairs $N$}\\
\cmidrule(l){2-8}
mechanism & 8 & 16 & 32 & 64 & 128 & 256 & 512\\
\midrule
softmax                      & 1.000 & 1.000 & 1.000 & 1.000 & 1.000 & 1.000 & 1.000\\
sliding-window softmax       & 0.003 & 0.002 & 0.002 & 0.002 & 0.002 & 0.002 & 0.002\\
linear attention             & 1.000 & 1.000 & 1.000 & 1.000 & 0.993 & 0.652 & 0.340\\
selective SSM (Mamba)        & 1.000 & 0.991 & 0.991 & 0.002 & 0.002 & 0.002 & 0.001\\
\midrule
direct access attention      & 1.000 & 1.000 & 1.000 & 1.000 & 1.000 & 1.000 & 1.000\\
\bottomrule
\end{tabular}
\end{table}

\begin{table}[t]
\centering\small
\caption{Decode FLOPs on the attention route to emit one token, for the same five mechanisms
and the same task sizes as Table~\ref{tab:main}, from \texttt{flops.py}. The context grows
with the task, $T{=}2N{+}80$, so softmax grows with it while the other four do not. Counts
the attention route only: the MLP and the projections are identical across mechanisms and
would swamp the comparison. Parameters are equal in the four attention-style rows; the SSM
row carries $9\%$ fewer, as Table~\ref{tab:main} notes.}
\label{tab:mainflops}
\setlength{\tabcolsep}{3.4pt}\begin{tabular}{lccccccc}
\toprule
& \multicolumn{7}{c}{number of pairs $N$}\\
\cmidrule(l){2-8}
mechanism & 8 & 16 & 32 & 64 & 128 & 256 & 512\\
\midrule
softmax                      & $185{,}856$ & $216{,}832$ & $278{,}784$ & $402{,}688$ & $650{,}496$ & $1{,}146{,}112$ & $2{,}137{,}344$\\
sliding-window softmax       & $15{,}488$ & $15{,}488$ & $15{,}488$ & $15{,}488$ & $15{,}488$ & $15{,}488$ & $15{,}488$\\
linear attention             & $16{,}896$ & $16{,}896$ & $16{,}896$ & $16{,}896$ & $16{,}896$ & $16{,}896$ & $16{,}896$\\
selective SSM (Mamba)        & $54{,}912$ & $54{,}912$ & $54{,}912$ & $54{,}912$ & $54{,}912$ & $54{,}912$ & $54{,}912$\\
\midrule
direct access attention       & $\textbf{1{,}694}$ & $\textbf{1{,}694}$ & $\textbf{1{,}694}$ & $\textbf{1{,}694}$ & $\textbf{1{,}694}$ & $\textbf{1{,}694}$ & $\textbf{1{,}694}$\\
\bottomrule
\end{tabular}
\end{table}

\section{The commitment term}\label{app:commit}

Two properties of the commitment term \eqref{eq:entloss} are measured here: that its weight
does not have to be tuned, and that its schedule does. Both are measured at $\dk{=}6$ with
eight heads and $N{=}128$. That setting gives $64$ cells per head against a context of $336$
written positions, which puts the memory well inside the regime where many keys share a cell
and commitment is therefore hard to obtain.

\paragraph{The weight does not have to be tuned.} Varying $\beta$ over two orders of magnitude
leaves the result unchanged: compiled accuracy is $0.998$, $0.998$, $0.998$, $0.998$ and
$0.997$ at $\beta=0.3,1,3,10,30$, with address concentration $1-H/\dk\ln2$ reaching $1.000$
in every case. Any $\beta$ in that range commits the address fully, and the value $1$ used
throughout is not critical within it.

\paragraph{The schedule is necessary.} The term supplies commitment, and only the task error
can supply coincidence, so the term cannot be activated before the task error has aligned the
codes. Activated from step $0$, the model does not learn the task. Its differentiable
read finishes at $0.002$, exactly the chance rate, with address concentration
at $1.000$ and query-key coincidence at $0.018$, close to the $1/64=0.016$ that random
agreement over $2^{6}$ cells would give. The addresses are fully committed, but to cells unrelated to the task, and because every bit
is saturated no gradient remains that could move them. Activated after the differentiable read
has converged, the precise activation step makes little difference:
activating it at $9{,}000$ and at $18{,}000$ both give $0.998$. This is why \eqref{eq:entloss} is scheduled on a
measured convergence rather than a step count, which would have to be retuned whenever the
task, the context length or the step budget changed.

\paragraph{The effect of delayed activation.} The term is activated on a measured accuracy, so what
matters is the behaviour of a run before that accuracy is reached. At $N{=}512$ with $22$ heads
(\texttt{main\_addr\_N512}) the key-code magnitude $|z_k|$ \emph{falls} from $0.794$ to
$0.547$ over the first $7{,}500$ steps while the address entropy rises from $5.56$ to $6.55$
nats, $79\%$ of the $8.32$-nat maximum at $\dk{=}12$: the addresses spread before the codes
align. Only at the step-$10{,}000$ evaluation does the differentiable read leave the
floor, reaching $1.000$. The term is activated there, and by the next evaluation $|z_k|$ has climbed
past $12$ and the address entropy has reached $0$. Key-query alignment develops during that
diffuse phase, and activating the term early prevents it.

\end{document}